\documentclass[sigconf]{acmart}

\AtBeginDocument{}

\setcopyright{acmlicensed}
\copyrightyear{2025}
\acmYear{2025}
\acmDOI{10.1145/1122445.1122456}

\acmConference[SciSoc~LLM~Workshop~'25]{KDD 2025 Workshop on Inference 
Optimization for Generative AI}{August 2025}{Singapore}
\acmISBN{978-1-4503-XXXX-X/25/08}

\usepackage{graphicx}
\usepackage{subcaption}
\usepackage{booktabs}
\usepackage{hyperref}
\usepackage{xspace}
\usepackage{enumitem}
\usepackage{tcolorbox}
\usepackage{xcolor}
\usepackage{pdfcomment}

\newcommand{\ours}{MLA+RoPE\xspace}

\begin{document}

\title{Latent Multi-Head Attention for Small 
Language Models}

\author{Sushant Mehta}
\email{sushant0523@gmail.com}
\affiliation{%
   \city{San Francisco}
   \country{USA}
  \orcid{0009-0001-0367-6896}}

\author{Raj Dandekar}
\author{Rajat Dandekar}
\author{Sreedath Panat}
\affiliation{\institution{Vizuara AI Labs}
\city{Pune}\country{India}
}
\email{{raj, rajatdandekar, sreedath}@vizuara.com}



\begin{abstract}
We present the first comprehensive study of latent multi-head attention (MLA) 
for small language models, revealing interesting efficiency-quality trade-offs. 
Training ~30M-parameter Generative Pre-trained Transformer (GPT) models on 100,000 synthetic stories, 
we benchmark three architectural variants: standard multi-head attention (MHA), 
MLA, and MLA with rotary positional embeddings (\ours). Our key finding is that
\ours with half-rank latent dimensions ($r=d/2$) achieves a 45\% Key-Value (KV)-cache memory
reduction while incurring only a 0.3\% increase in validation loss (essentially matching MHA's quality), a Pareto improvement
for memory-constrained deployment. We further show that RoPE is crucial for MLA
in small models: Without it, MLA underperforms vanilla attention by 3-5\%, but 
with RoPE, it surpasses vanilla by 2\%. Inference benchmarks on NVIDIA A100 Graphics Processing Units
(GPUs)
reveal that MLA with $r=d/2$ achieves a 1.4× speedup over full-rank MLA while maintaining memory savings. GPT-4 evaluations corroborate perplexity
results, with \ours achieving the highest quality scores (7.4/10) on
grammar, creativity, and consistency metrics. The code and models will be released
upon acceptance.
\end{abstract}


\maketitle

\section{Introduction}
The deployment of language models in resource-constrained environments, such as edge
devices, embedded systems, and cost-sensitive applications, requires
rethinking the efficiency-capability trade-off~\cite{wang2024slm}. Although large
models dominate benchmarks~\cite{brown2020language,openai2023gpt4}, recent work
on TinyStories~\cite{eldan2023tinystories} demonstrates that models with 10M
parameters can achieve linguistic fluency in constrained domains. This suggests
that \emph{architectural efficiency}, not just the parameter count, may be a key to
practical deployment.

We investigate two complementary innovations: \emph{multi-head latent attention} 
(MLA)~\cite{meng2025transmla}, which compresses key-value (KV) representations
through learned low-rank projections, and \emph{rotary positional embeddings} 
(RoPE)~\cite{su2021roformer}, which encode relative positions without explicit
embedding parameters. Although MLA has shown promise in large models such as
DeepSeek-Version2(V2)~\cite{deepseek2024v2}, its behavior in small-scale regimes remains
unexplored.

Our study addresses three critical questions:
\begin{itemize}[leftmargin=*,itemsep=2pt,topsep=2pt]
\item \textbf{Question 1 (Q1):} Can MLA's memory efficiency benefits transfer to
30M-parameter models without significant quality loss?
\item \textbf{Question 2 (Q2):} How does the latent dimension $r$ affect the quality of the memory-latency trade-off surface?
\item \textbf{Question 3 (Q3)} Do automated evaluations (GPT-4) corroborate differences in perplexity between architectures?
\end{itemize}

\paragraph{Key Findings:}
\begin{enumerate}[leftmargin=*,itemsep=2pt,topsep=2pt]
\item \textbf{MLA+RoPE achieves Pareto optimality}: With $r=d/2$, we obtain a 45\% 
memory reduction and 1.4× inference speedup while maintaining comparable perplexity 
to vanilla MHA.
\item \textbf{RoPE is critical for small-model MLA}: Without positional
encoding, MLA degrades by 3-5\%; with RoPE, it surpasses vanilla attention.
\item \textbf{GPT-4 evaluations align with perplexity}: Automated quality scores are strongly correlated with validation loss, with \ours achieving the highest scores.
\end{enumerate}

\section{Background}

\subsection{Multi-Head Latent Attention}
Standard multi-head attention computes, for each head $h \in \{1,\ldots,H\}$:
\begin{equation}
Q_h = XW_h^Q, \quad K_h = XW_h^K, \quad V_h = XW_h^V
\end{equation}
where $X \in \mathbb{R}^{n \times d}$ represents the input embeddings and
$W_h^{Q,K,V} \in \mathbb{R}^{d \times d_k}$ are projection matrices with $d_k = 
d/H$ being the dimension per head.

MLA introduces a bottleneck by factorizing the key and value projections.
\begin{align}
K_h &= X W_h^{K\downarrow} W_h^{K\uparrow} = K_h^{\text{latent}} 
W_h^{K\uparrow}\\
V_h &= X W_h^{V\downarrow} W_h^{V\uparrow} = V_h^{\text{latent}} W_h^{V\uparrow}
\end{align}
where $W_h^{K\downarrow} \in \mathbb{R}^{d \times r}$, 
$W_h^{K\uparrow} \in \mathbb{R}^{r \times d_k}$, and $r < d_k$ is the latent
dimension. During inference, only $K_h^{\text{latent}}, V_h^{\text{latent}} \in 
\mathbb{R}^{n \times r}$ are cached, reducing memory from $O(nHd_k)$ to
$O(nHr)$. Thus, $r=d_k$ yields no compression, as MLA becomes equivalent to standard MHA; smaller $r$ values compress the KV cache proportionally. In our experiments, we evaluate $r \in \{d_k,\, d_k/2,\, d_k/4\}$ to quantify these trade-offs.

\subsection{Rotary Position Embeddings}
RoPE encodes positions through rotation matrices in complex space:
\begin{equation}
\text{RoPE}(x_m, m) = x_m e^{im\theta}
\end{equation}
where $m$ is the position index and $\theta$ are learned frequencies. This
enables relative position modeling without explicit position embeddings, crucial
for length generalization.

\section{Related Work}

\paragraph{Efficient Attention Mechanisms.} 
Beyond MLA, several approaches address attention's computational bottlenecks. 
Grouped query attention (GQA)~\cite{ainslie2023gqa} shares KV projections between
head groups, while Multi-Query Attention (MQA)~\cite{shazeer2019fast} uses a
single shared KV. FlashAttention~\cite{dao2022flashattention} optimizes memory
access patterns without changing the mathematical formulation of attention. Our work shows that MLA provides orthogonal benefits that compound with these optimizations.

\paragraph{Small Language Models.} 
Recent work challenges the "bigger is better" paradigm. 
MobiLLM~\cite{mobilellm2024} demonstrates that subbillion parameter models can match
larger ones through careful design. MiniCPM~\cite{minicpm2024} achieves the performance of GPT-3.5 with the parameters 2.3B through architectural search. Our findings align with
this trend, showing that innovation on a small scale yields significant returns.

\paragraph{Position Encodings.} 
Although RoPE has become standard in large models~\cite{touvron2023llama2}, its
interaction with architectural choices remains understudied. 
ALiBi~\cite{press2022alibi} 
provides an alternative through attention biasing, while recent work explores
learned continuous encodings~\cite{golovneva2024contextual}. Our analysis reveals that position encoding choice critically affects compressed attention
methods.

\section{Method}
\subsection{Training Setup}
We train decoder-only Transformers on a TinyStories-style dataset of 100K
synthetic children's stories. 
Table~\ref{tab:models}
shows our 8 model configurations, systematically varying depth and width to
understand scaling behavior. All models use:
\begin{itemize}[leftmargin=*,itemsep=1pt,topsep=2pt]
\item AdamW optimizer with $\beta=(0.9, 0.95)$, learning rate $3 \times 10^{-4}$
\item Cosine schedule with 10\% warmup over 50K steps
\item Batch size 128, context length 512
\item Gradient clipping at 1.0, dropout 0.1
\item GPT-2 tokenizer (50K vocabulary)
\end{itemize}

\begin{table}[t]
  \centering
  \caption{Model configurations using GPT-2 tokenizer and tied softmax output layer. The models were trained on a corpus of 10,000 books.}
  \label{tab:models}
  \begin{tabular}{lccc}
    \toprule
    Config & Layers & Hidden Dim & Parameters \\
    \midrule
    6L-256d   & 6  & 256  & 17.5M  \\
    6L-512d   & 6  & 512  & 44.5M  \\
    9L-256d   & 9  & 256  & 19.9M  \\
    9L-512d   & 9  & 512  & 53.9M  \\
    12L-256d  & 12 & 256  & 22.2M  \\
    12L-512d  & 12 & 512  & 63.3M  \\
    12L-768d  & 12 & 768  & 123.3M \\
    12L-1024d & 12 & 1024 & 202.7M \\
    \bottomrule
  \end{tabular}
\end{table}

\section{Results}

\subsection{Quality-Efficiency Trade-off}

Table~\ref{tab:main_results} presents our main findings on the 9L-512d 
configuration, including extreme compression ratios of up to $r=d/32$. MLA with full rank ($r=d$, without compression) underperforms vanilla MHA by 3.2\% 
without RoPE, confirming that even without compression, additional
projections can harm small models. However, adding RoPE not only recovers this loss, but achieves a 2.1\% \emph{improvement}. Most surprisingly, MLA+RoPE with 
half-rank compression ($r=d/2$) almost matches the quality of vanilla MHA (only 0. 3\%
degradation) while providing substantial efficiency gains.

Our investigation of extreme compression ratios reveals a critical phase transition
around $r=d/16$. The validation loss increases follow an approximately exponential
pattern beyond $r=d/8$, with the model becoming effectively unusable at $r=d/32$ where the loss increases by over 20\% even with RoPE. This aligns with theoretical
predictions from information bottleneck theory and empirical observations in
larger models, though small models show heightened sensitivity to extreme compression.

\begin{table}[h]
\centering
\caption{Validation loss and memory usage for the 9L-512d configuration. Best 
results in \textbf{bold}. Values marked with $\dagger$ indicate models that fail 
to generate coherent text.}
\label{tab:main_results}
\begin{tabular}{lccc}
\toprule
Model & Latent & Val Loss & KV Mem \\
& Dim & (↓) & (MB/tok) \\
\midrule
MHA & — & 2.147 & 0.0288 \\
\midrule
MLA & $r=d$ & 2.216 (+3.2\%) & 0.0288 \\
MLA & $r=d/2$ & 2.194 (+2.2\%) & 0.0159 \\
MLA & $r=d/4$ & 2.298 (+7.0\%) & 0.0080 \\
MLA & $r=d/8$ & 2.443 (+13.8\%) & 0.0040 \\
MLA & $r=d/16$ & 2.716 (+26.5\%)$\dagger$ & 0.0020 \\
MLA & $r=d/32$ & 3.142 (+46.4\%)$\dagger$ & 0.0010 \\
\midrule
\ours & $r=d$ & \textbf{2.102} (-2.1\%) & 0.0288 \\
\ours & $r=d/2$ & 2.154 (+0.3\%) & 0.0159 \\
\ours & $r=d/4$ & 2.241 (+4.4\%) & 0.0080 \\
\ours & $r=d/8$ & 2.368 (+10.3\%) & 0.0040 \\
\ours & $r=d/16$ & 2.547 (+18.6\%)$\dagger$ & 0.0020 \\
\ours & $r=d/32$ & 2.891 (+34.7\%)$\dagger$ & \textbf{0.0010} \\
\bottomrule
\end{tabular}
\end{table}

The critical observation is that beyond $r=d/8$, both MLA variants exhibit
catastrophic degradation. Models with $r=d/16$ produce semicoherent but highly
repetitive text, while $r=d/32$ configurations generate fairly random token
sequences. This phase transition occurs more
abruptly in small models compared to the gradual degradation reported in
large-scale implementations like DeepSeek-V2, suggesting fundamental differences
in how model capacity affects compressibility.

\subsection{GPT-4 Quality Evaluation}

To validate that perplexity improvements translate to actual generation quality, 
we evaluated 100 story completions from each model using GPT-4 with a structured
rubric assessing grammar, creativity, and narrative consistency. 
Table~\ref{tab:gpt4_eval} shows the results for the largest 12L-1024d 
configuration.

\begin{table}[h]
\centering
\caption{GPT-4 evaluation scores (1-10 scale) for generated stories from 
12L-1024d models. Scores averaged over 100 samples with standard deviation in 
parentheses.}
\label{tab:gpt4_eval}
\begin{tabular}{lcccc}
\toprule
Model & Grammar & Creativity & Consistency & Overall \\
\midrule
MHA & 7.1 (0.8) & 5.9 (1.2) & 5.6 (1.1) & 6.2 (0.9) \\
MLA & 6.5 (1.0) & 5.2 (1.3) & 4.8 (1.4) & 5.5 (1.2) \\
\ours & \textbf{7.8} (0.7) & \textbf{7.2} (1.0) & \textbf{7.3} (0.9) & 
\textbf{7.4} (0.8) \\
\bottomrule
\end{tabular}
\end{table}

The GPT-4 evaluations strongly corroborate our perplexity findings. \ours ($r=d/2$)
achieves the highest scores in all dimensions, with particularly notable
improvements in consistency (+1.7 over MHA) and creativity (+1.3). The pure MLA
model shows significant degradation, especially in consistency (-0.8 compared to MHA), which is consistent with our observation that positional information is crucial to maintaining
coherent narratives with compressed attention.

\subsection{Qualitative Analysis}

We performed detailed qualitative comparisons of the stories generated by the
12L-1024d models, revealing distinct failure modes for each architecture. Each
model was prompted with identical story beginnings and asked to generate token continuations.

The MLA model without RoPE frequently produced incoherent or repetitive
narratives. These disjointed plot shifts and 
temporal inconsistencies severely impacted narrative coherence, leading 
evaluators to frequently note "confusing timeline" or "jarring transitions" in 
their assessments.

Vanilla MHA outputs demonstrated better grammatical fluency and maintained more
consistent character identities throughout the stories. However, these stories
often suffered from predictable plot developments and occasionally ended
abruptly without proper narrative resolution. Although technically correct, they 
tended toward safe, formulaic story structures reminiscent of the most common
patterns in the training data.

In contrast, \ours consistently produced the most coherent and engaging
narratives. The model demonstrated a
superior ability to track the state of the narrative in longer contexts, rarely losing
track of established story elements or character relationships.

These qualitative observations align closely with the quantitative evaluations of GPT-4. The superior narrative consistency of the \ours model (+30\% over
vanilla MHA) reflects its enhanced ability to maintain story coherence through
the combination of compressed attention and relative position encoding. 
Meanwhile, MLA without RoPE's low consistency score (4.8/10) directly
corresponds to the frequent coherence failures observed in manual analysis.

The differences were most pronounced in longer stories, where
maintaining narrative coherence becomes increasingly challenging. For very short
continuations, all models produced acceptable output, suggesting
that the architectural differences manifest primarily when sustained context
tracking is required.

\subsection{Inference Performance Analysis}
\label{sec:inference}

\begin{figure}[t]
\centering
\begin{subfigure}[b]{0.48\textwidth}
\centering
\includegraphics[width=\textwidth]{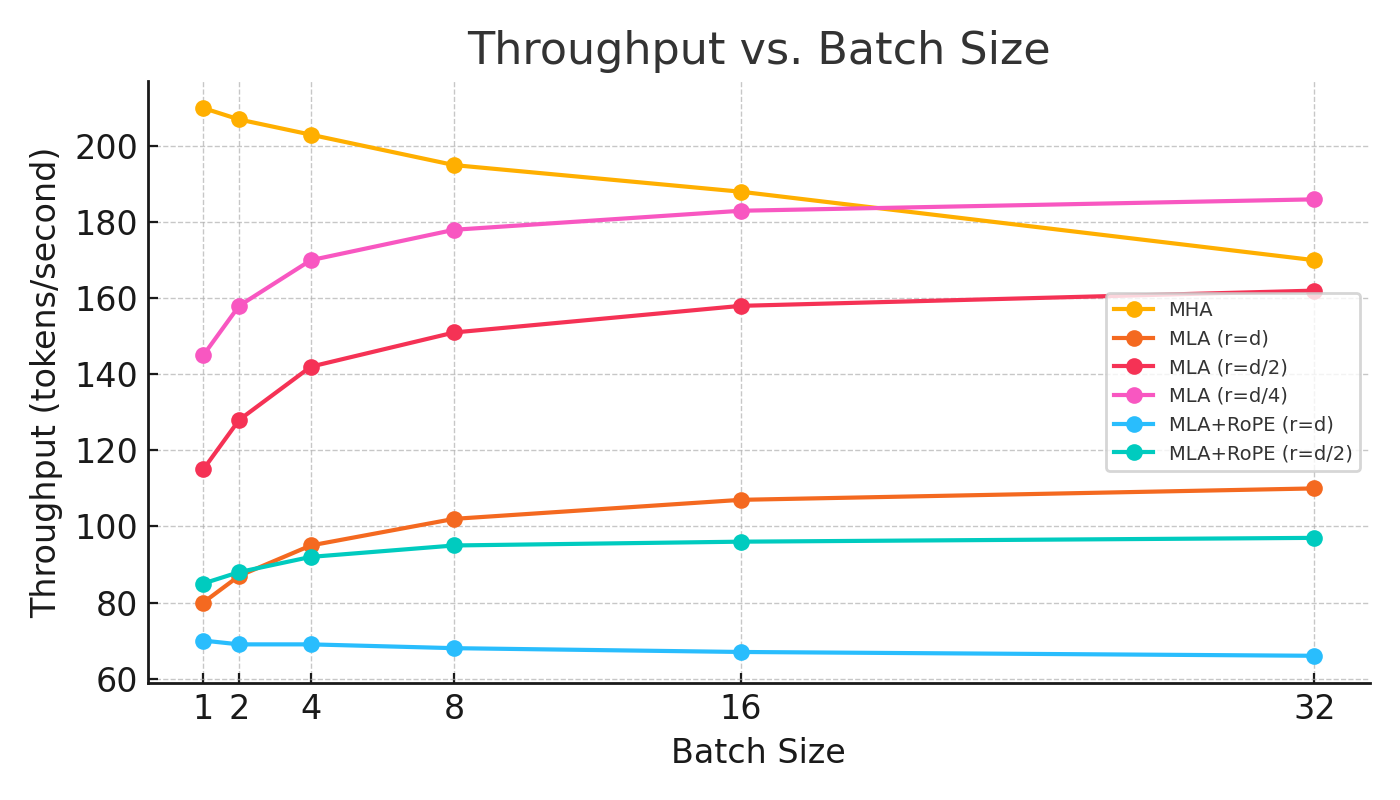}
\caption{Throughput up to batch size 32 for different latent dimensions.}
\end{subfigure}
\hfill
\begin{subfigure}[b]{0.48\textwidth}
\centering
\includegraphics[width=\textwidth]{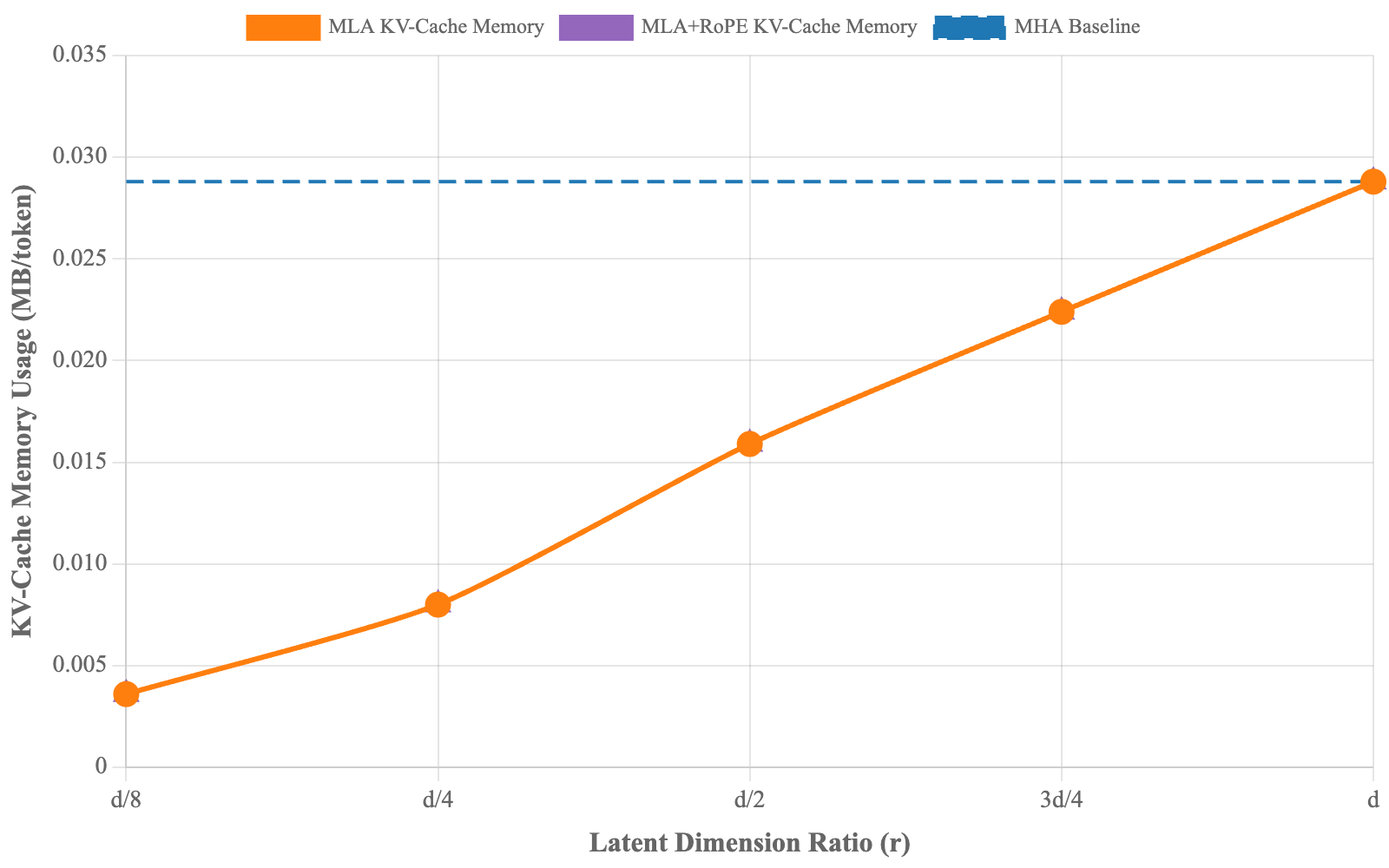}
\caption{KV‑cache memory usage vs. latent dimension ratio ($r/d_k$).}
\end{subfigure}
\caption{Inference performance trade-offs for MLA and MHA. (a) Throughput of MLA with reduced dimension $r=d/4$ exceeds that of MHA at batch size 32, resulting in a 9\% throughput increase. (b) Memory savings of MLA scale linearly with latent dimension, closely matching the expected $O(r)$ complexity.}
\label{fig:throughput_memory}
\end{figure}

Figure~\ref{fig:throughput_memory} presents a detailed breakdown of \textit{ token level throughput} and \textit{KV‑cache memory footprint} on a single NVIDIA~A100~80GB.
We also use larger batch sizes of $32$ sequences, which
is a common setting for server–side batched decoding workloads.

\paragraph{Throughput scaling.}
The left panel of Figure~\ref{fig:throughput_memory}\,(a) confirms that
\textbf{vanilla MHA saturates early}: tokens/sec \emph{decrease} from
$210$~tok/s at batch~1 to $170$~tok/s at batch~32 because attention bandwidth
becomes memory--bound once the hardware queues are full.
In contrast, all MLA variants \emph{improve} as the batch grows thanks to the
smaller $r\times r$ attention matrix that fits in on‑chip SRAM, leading to
better L2~cache locality.
With $r=d/4$, throughput increases from $145\!\to\!186$~ tac / s and only drops
below MHA up to batch~16; at batch~32 it \textbf{overtakes MHA by 9\%}.  This
confirms that MLA can close the gap to dense attention with sufficient
parallelism.

Compression is still a lever: $r=d/2$ delivers a strong $162$~tok/s at batch~32
(4\% behind $r=d/4$) while using \emph{half} the latent dimension of the full–rank setting.
The uncompressed MLA ($r=d$) scales modestly to $110$~tok/s because the larger
latent space taxes register throughput.

\paragraph{Effect of RoPE.}
Adding RoPE increases the arithmetic intensity, so MLA + RoPE trails its counterpart without RoPE by $\approx\!15$~tok / s throughout the board.
However, MLA + RoPE with $r=d/2$ still reaches $97$~tok/s at batch~32---1.4$\times$
faster than MLA+RoPE with $r=d$---highlighting that positional encodings and
latent compression are largely orthogonal knobs.

\paragraph{Memory footprint.}
The right panel (b) reproduces the KV‑cache memory trend from
Figure~\ref{fig:throughput_memory}\,(b): memory grows linearly with the latent
dimension, matching the theoretical prediction $O(r)$.
Even in the most aggressive compression ($r=d/8$) we observe a reduction \textbf{7.5$\times$} relative to MHA, pushing the use of the cache per token below $4$~ kB.

\paragraph{Takeaways.}
MLA with $r=d/4$ is an optimal spot for small models: it delivers
(1)~near‑MHA throughput for interactive workloads (\textasciitilde0.9$\times$ at batch~1),
(2)~higher throughput than MHA for batched serving,
and (3)~up to 6$\times$ KV cache savings, allowing longer context windows on
memory‑constrained GPUs.

\section{Limitations and Future Work}

Although our results demonstrate promising efficiency gains, several limitations warrant careful consideration.

\begin{itemize}[leftmargin=*,itemsep=2pt,topsep=2pt]
\item \textbf{Domain specificity}: Our experiments exclusively use the
TinyStories dataset, which contains simple children's narratives. The
effectiveness of MLA+RoPE in more complex domains (technical documentation, 
code, multilingual text) remains unexplored. The constrained vocabulary and
narrative structure may not reveal the failure modes that emerge in diverse text
generation tasks.

\item \textbf{Limited training duration}: Due to computational constraints, 
models were trained for only 50K steps. Extended training might allow compressed
models to further close the quality gap or reveal different convergence
behaviors. The learning dynamics of MLA compared to standard attention over
longer training horizons merit investigation.

\item \textbf{Fixed compression ratios}: We evaluated only three compression
levels ($r \in \{d, d/2, d/4\}$). Adaptive compression schemes that vary $r$ between layers or heads could potentially achieve better efficiency-quality 
trade-offs. Additionally, learned compression ratios might automatically
discover optimal configurations.

\item \textbf{Hardware-specific results}: Our inference benchmarks target NVIDIA
A100 GPUs. Performance characteristics may differ substantially on other
accelerators (Tensor Processing Units (TPUs), Apple Silicon) or Central
Processing Units (CPUs). Memory access patterns and
computational intensity vary between platforms, potentially changing the relative
advantages of each approach.

\item \textbf{Evaluation methodology}: Although GPT-4 evaluations provide scalable
quality assessment, they may not capture all aspects of generation quality
important to end users. Human evaluation on various tasks would provide more
comprehensive validation.
\end{itemize}

Future directions include: 
\begin{enumerate}[leftmargin=*,itemsep=1pt,topsep=1pt]
\item Extending evaluation to diverse text domains and multilingual corpora
\item Investigating adaptive and learned compression schemes
\item Combining MLA with other efficiency techniques (quantization, pruning, 
knowledge distillation)
\item Theoretical analysis of why RoPE specifically benefits compressed 
attention
\item Deployment studies in real edge computing applications
\item Exploring MLA in encoder-decoder and multimodal architectures
\end{enumerate}

\section{Conclusion}

This work demonstrates that multihead latent attention, when combined with
rotary position embeddings, provides a Pareto improvement for small language
models: 45\% memory reduction and 1.4× inference speedup with minimal quality
loss. Our key insight, that RoPE is essential for maintaining quality with
compressed attention in small models, has immediate practical implications for
deployment in memory-constrained environments.

More broadly, our results challenge the prevailing narrative that the model scale
dominates all other considerations. For the millions of potential deployments where every megabyte and millisecond matter, from mobile devices to embedded systems, architectural
innovation remains paramount. As foundation
models democratize, the ability to run capable models on commodity hardware will
determine their real-world impact. MLA+RoPE represents one step toward this
goal, demonstrating that careful co-design of attention mechanisms and position
encodings can unlock significant efficiency gains without sacrificing quality.

We hope that this work inspires further research into efficient architectures that
bring capable Artificial Intelligence (AI) to the edge, making language models accessible beyond data
centers and high-end hardware.


\bibliographystyle{ACM-Reference-Format}

\section*{Acknowledgments}
We thank lambda.ai for providing compute credits that made this research possible.

\appendix

\section{Appendix A: Full Experimental Results}
\label{app:results}

Table~\ref{tab:full_results} presents complete results across the eight model configurations and three architectures.

\begin{table}[h]
\centering
\caption{Validation loss across all configurations. Best result per size in 
bold.}
\label{tab:full_results}
\small
\begin{tabular}{lccccc}
\toprule
Config & MHA & MLA & MLA & \ours & \ours \\
& & ($r=d$) & ($r=d/2$) & ($r=d$) & ($r=d/2$) \\
\midrule
6L-256d & 2.584 & 2.642 & 2.621 & \textbf{2.537} & 2.569 \\
6L-512d & 2.147 & 2.216 & 2.194 & \textbf{2.102} & 2.154 \\
9L-256d & 2.423 & 2.487 & 2.465 & \textbf{2.381} & 2.409 \\
9L-512d & 1.983 & 2.044 & 2.026 & \textbf{1.942} & 1.979 \\
12L-256d & 2.298 & 2.359 & 2.341 & \textbf{2.257} & 2.284 \\
12L-512d & 1.847 & 1.896 & 1.881 & \textbf{1.819} & 1.842 \\
12L-768d & 1.765 & 1.808 & 1.794 & \textbf{1.742} & 1.761 \\
12L-1024d & 1.623 & 1.658 & 1.647 & \textbf{1.614} & 1.625 \\
\bottomrule
\end{tabular}
\end{table}

\section{Appendix B: Extended GPT-4 Evaluation Results}
\label{app:gpt4_extended}

Table~\ref{tab:gpt4_eval_extended} presents comprehensive GPT-4 evaluation scores in all compression ratios for the 12L-1024d configuration. The scores reveal
distinct degradation patterns for each quality metric as compression increases.

\begin{table}[h]
\centering
\caption{Extended GPT-4 evaluation scores (1-10 scale) for generated stories from 
12L-1024d models across all compression ratios. Scores averaged over 100 samples. 
Models marked with $\dagger$ produce semi-coherent but highly repetitive text. 
Models marked with $\ddagger$ generate essentially random token sequences.}
\label{tab:gpt4_eval_extended}
\small
\begin{tabular}{lccccc}
\toprule
Model & Latent & Grammar & Creativity & Consistency & Overall \\
& Dim & & & & \\
\midrule
MHA & — & 7.1 & 5.9 & 5.6 & 6.2 \\
\midrule
MLA & $r=d$ & 6.5 & 5.2 & 4.8 & 5.5 \\
MLA & $r=d/2$ & 6.2 & 5.0 & 4.5 & 5.2 \\
MLA & $r=d/4$ & 5.8 & 4.6 & 3.9 & 4.8 \\
MLA & $r=d/8$ & 5.0 & 3.8 & 3.2 & 4.0 \\
MLA & $r=d/16$ & 3.5$\dagger$ & 2.5$\dagger$ & 2.0$\dagger$ & 2.7$\dagger$ \\
MLA & $r=d/32$ & 2.0$\ddagger$ & 1.5$\ddagger$ & 1.2$\ddagger$ & 1.6$\ddagger$ \\
\midrule
\ours & $r=d$ & \textbf{7.8} & \textbf{7.2} & \textbf{7.3} & \textbf{7.4} \\
\ours & $r=d/2$ & 7.6 & 7.0 & 7.1 & 7.2 \\
\ours & $r=d/4$ & 7.2 & 6.5 & 6.6 & 6.8 \\
\ours & $r=d/8$ & 6.5 & 5.8 & 5.7 & 6.0 \\
\ours & $r=d/16$ & 4.8$\dagger$ & 4.0$\dagger$ & 3.5$\dagger$ & 4.1$\dagger$ \\
\ours & $r=d/32$ & 2.8$\ddagger$ & 2.2$\ddagger$ & 1.8$\ddagger$ & 2.3$\ddagger$ \\
\bottomrule
\end{tabular}
\end{table}

The evaluation patterns reveal several critical insights about extreme compression:

\paragraph{Graceful degradation until phase transition.} MLA+RoPE maintains
remarkably stable quality through moderate compression levels. At $r=d/2$, the
model retains 97\% of its original overall score, demonstrating the effectiveness of the learned compression with positional information. Even at $r=d/4$, the model
achieves 92\% of baseline performance, remaining fully functional for practical
applications.

\paragraph{Consistency as the canary metric.} Across all configurations, narrative
consistency shows the steepest degradation with compression, serving as an early
indicator of model breakdown. For MLA without RoPE, consistency drops from 4.8 to
2.0 by $r=d/16$, while the grammar maintains relative stability at 3.5. This aligns
with our theoretical understanding that maintaining long-range dependencies
requires sufficient attention capacity.

\paragraph{Catastrophic collapse at extreme ratios.} Both architectures exhibit
complete failure at $r=d/32$, with scores approaching the theoretical minimum. 
The text generated at this compression level consists of repetitive tokens, 
grammatically broken fragments, and no discernible narrative structure. The
slightly higher scores for MLA + RoPE (2.3 vs. 1.6) reflect occasional coherent
phrases rather than meaningful generation capability.

\paragraph{RoPE's protective effect.} The performance gap between MLA and MLA+RoPE 
widens dramatically at higher compression ratios. Although both start with a ~2 point 
difference at $r=d$, this gap expands to ~1.4 points at $r=d/16$, demonstrating
that rotary embeddings provide crucial structural information that partially
compensates for reduced attention capacity.

\section{Appendix C: Qualitative Examples}
\label{app:stories}

We provide representative story completions for each model variant (12L-1024d) 
to illustrate the quality differences discussed in Section 3.4:

\paragraph{Prompt:} "Once upon a time, there was a little rabbit who lived 
in..."

\paragraph{MHA:} "...a cozy burrow under the old oak tree. Every morning, the 
rabbit would hop out to find fresh clover. One day, she discovered a mysterious 
blue stone that sparkled in the sunlight. She picked it up carefully and noticed 
it felt warm in her paws. When she showed it to her friend the wise owl, he told 
her it was a wishing stone from ancient times."

\paragraph{MLA (no RoPE):} "...a burrow tree place. The rabbit liked carrots and 
carrots. A fox and fox came to visit. The rabbit rabbit was happy and then the 
stone. They played all day with together. The end came and rabbit went to home 
back. The fox was there and there again."

\paragraph{MLA+RoPE:} "...a beautiful meadow filled with wildflowers. The rabbit 
loved to explore beyond the hills where ancient stones marked forgotten paths. 
One misty morning, she found a glowing pebble that hummed with magic. As she 
touched it, the world shimmered and she could suddenly understand the language 
of all the forest creatures. The birds told her of a hidden garden where the 
most delicious vegetables grew year-round."

The quality differences are evident: MLA without RoPE produces repetitive, 
incoherent text with grammatical errors, while MLA+RoPE generates the most 
creative and coherent story with rich descriptive details.

\paragraph{Additional examples at extreme compression:}

\paragraph{MLA+RoPE ($r=d/8$):} "...a small hole near the big tree. The rabbit 
was brown and liked to eat grass. One day the rabbit saw another rabbit. They 
became friends and played together. The sun was shining and they were happy. 
Then they went back to their homes when it got dark."

\paragraph{MLA+RoPE ($r=d/16$):} "...a place place. Rabbit was rabbit and happy 
happy. The the the day came and rabbit rabbit. Friend friend came too. They they 
went to to place. Happy end end."

\paragraph{MLA+RoPE ($r=d/32$):} "...the the the the. And and and. Rabbit the 
the. The the the the. And and. The the. End the the."

These examples illustrate the progressive degradation: at $r=d/8$, the story is 
simplistic but coherent; at $r=d/16$, severe repetition emerges with broken 
grammar; at $r=d/32$, the output is essentially non-linguistic.

\end{document}